\documentclass[10pt, a4paper]{article}
\pdfoutput=1
\usepackage{lrec2022} 
\usepackage{multibib}
\newcites{languageresource}{Language Resources}
\usepackage{graphicx}
\usepackage{tabularx}
\usepackage{soul}
\usepackage{titlesec}
\titleformat{\section}{\normalfont\large\bfseries\center}{\thesection.}{1em}{}
\titleformat{\subsection}{\normalfont\SmallTitleFont\bfseries\raggedright}{\thesubsection.}{1em}{}
\titleformat{\subsubsection}{\normalfont\normalsize\bfseries\raggedright}{\thesubsubsection.}{1em}{}
\renewcommand\thesection{\arabic{section}}
\renewcommand\thesubsection{\thesection.\arabic{subsection}}
\renewcommand\thesubsubsection{\thesubsection.\arabic{subsubsection}}

\usepackage{epstopdf}
\usepackage[utf8]{inputenc}

\usepackage{hyperref}
\usepackage{xstring}

\usepackage{color}

\title{ALBETO and DistilBETO: Lightweight Spanish Language Models}

 \name{Jos\'e Cañete\textsuperscript{1,3},
 Sebasti\'an Donoso\textsuperscript{1,3},
 Felipe Bravo-Marquez\textsuperscript{1,3,4},
 \\
 {\bf \large
  Andr\'es Carvallo\textsuperscript{2,3},
  Vladimir Araujo\textsuperscript{2,3,4,5}
 }}

 \address{\textsuperscript{1} Department of Computer Science, University of Chile, Santiago, Chile \\
 \textsuperscript{2}Pontificia Universidad Cat\'olica de Chile, Santiago, Chile \\
\textsuperscript{3} National Center for Artificial Intelligence (CENIA), Santiago, Chile \\
\textsuperscript{4} Millennium Institute for Foundational Research on Data (IMFD), Santiago, Chile \\
\textsuperscript{5}KU Leuven, Leuven, Belgium \\
 \href{mailto:jcanete@dcc.uchile.cl}{\texttt{jcanete@dcc.uchile.cl}}\\}

\abstract{
In recent years there have been considerable advances in pre-trained language models, where non-English language versions have also been made available. Due to their increasing use, many lightweight versions of these models (with reduced parameters) have also been released to speed up training and inference times. However, versions of these lighter models (e.g., ALBERT, DistilBERT) for languages other than English are still scarce. In this paper we present ALBETO and DistilBETO, which are versions of ALBERT and DistilBERT pre-trained exclusively on Spanish corpora. We train several versions of ALBETO ranging from 5M to 223M parameters and one of DistilBETO with 67M parameters.
We evaluate our models in the GLUES benchmark that includes various natural language understanding tasks in Spanish. The results show that our lightweight models achieve competitive results to those of BETO (Spanish-BERT) despite having fewer parameters. More specifically, our larger ALBETO model outperforms all other models on the MLDoc, PAWS-X, XNLI, MLQA, SQAC and XQuAD datasets. However, BETO remains unbeaten for POS and NER. As a further contribution, all models are publicly available to the community for future research. \\ \newline }

\begin{document}

\maketitleabstract

\section{Introduction}\label{sec:intro}

The area of Natural Language Processing has gained a lot of interest in recent years due to the great success of deep neural networks in different tasks such as sentiment analysis \cite{socher2013recursive,rosenthal2017semeval,mohammad2018semeval}, sentence \cite{warstadt2019neural} and document classification \cite{SCHWENK18.658,carvallo2020automatic}, natural language inference \cite{williams2017broad,conneau2018xnli}, among others. 

This success can be explained by a large number of trainable parameters of these models, a considerable amount of training data, and the processing capacity power available nowadays, allowing them to learn complex relationships and cast complex functions.

One of these highly successful models is BERT \cite{devlin-etal-2019-bert}.
This model, based on a Transformer \cite{vaswani2017attention} encoder, is pre-trained with lots of unlabeled text data and fine-tuned to specific tasks. 

BERT, in its large version, has 330 million parameters. Moreover, one of the most larger models is GPT-3 \cite{brown2020language} which has 175 billion parameters.

A problem with these large models is the difficulty of getting them into production on real-time applications (web services) or hardware-restricted devices, such as mobile devices, because of their high memory and computation requirements. Although there have been efforts for solving this issue proposing lighter models such as ALBERT \cite{Lan2020ALBERT}, DistilBERT \cite{sanh2020distilbert}, TinyBERT \cite{jiao2020tinybert}, or PKD-BERT \cite{sun-etal-2019-patient}, these versions of models have been trained exclusively in English.

On the other hand, there are recent efforts in the community to provide trained models in Spanish, such as Spanish-BERT (BETO) \cite{CaneteCFP2020} and Spanish-RoBERTa \cite{gutierrez2022maria,de2022bertin,perez2021robertuito}. However, there is still a lack of lighter and more efficient models for the Spanish language.

In this paper we present ``A Lite version of BETO'' (ALBETO) and a ``Distill version of BETO'' (DistilBETO) in order to democratize the use of these models in Spanish-speaking regions. We follow the same pre-training methodology as ALBERT and DistilBERT but with the difference that our models were trained exclusively with a corpus in Spanish.

We present an evaluation of all models on a variety of tasks and we also release the code\footnote{\url{https://github.com/dccuchile/lightweight-spanish-language-models}} used to fine-tune the models on the tasks as well as the fine-tuned models\footnote{\url{https://huggingface.co/dccuchile}}. 
Our results indicate that in specific tasks, namely Part of Speech (POS) and Named Entity Recognition (NER), having a lighter version of the language models implies a slight reduction in the performance compared to the standard version of BETO. However, the lightweight models' performance improves despite having fewer parameters in other tasks, such as Natural Language Inference, Paraphrase Identification, Document Classification and Question Answering.
In general, by fine-tuning our ALBETO models, we achieved 87\% of BETO performance while having 22 times fewer parameters in the case of ALBETO \textit{tiny} and 97\% of BETO performance at 9 times that size in the case of ALBETO \textit{base}.

The remainder of this article is structured as follows. In Section \ref{sec:rel}, we provide
a review of related work. In Section \ref{sec:data}, our models and the data on which they are trained are described. In Section \ref{sec:eval}, we present the experiments we conducted to evaluate the proposed resources and discuss results. Finally, in Section \ref{sec:conc} the main findings and conclusions are discussed.

\section{Related Work}\label{sec:rel}

\subsection{Multilingual and Monolingual BERT}

Multilingual models are models that are trained simultaneously using data from several languages. An example of these models is mBERT \cite{pires2019multilingual}, a version of the BERT model trained in English and 104 languages.  Interestingly, some studies  \cite{wu2019beto,DBLP:journals/corr/abs-1906-01502} have shown how these multi-language learners set strong baselines for non-English tasks.

Besides, several single language BERT models have been released. For instance, CamemBERT~\cite{martin2019camembert} and FlauBERT~\cite{le2019flaubert} for French, BERTje~\cite{de2019bertje} and RobBERT~\cite{delobelle2020robbert} for Dutch, FinBERT~\cite{virtanen2019multilingual} for Finish, among others. Those models have shown better performance than multilingual ones, demonstrating the importance of having them for language-specific tasks.

In the case of the Spanish language, we can find BETO \cite{CaneteCFP2020}, which is a BERT base model, and a family of RoBERTa models. RoBERTa-BNE \cite{gutierrez2022maria}, which is available in base and large versions was trained on the corpus crawled by the National Library of Spain. BERTIN \cite{de2022bertin} is a base sized model that was trained on the Spanish portion of mC4 \cite{xue2020mt5}. RoBERTuito \cite{perez2021robertuito} is a RoBERTa base model trained exclusively on tweets. However, there is a lack of more efficient, smaller, or lighter language models exclusively trained with Spanish data.

\subsection{Efficient Transformer-based Language Models}


Several architectures have been designed to avoid overcomputing in Transformer-based models. One such strategy is to use lightweight architectures that are trained from scratch. 
As an example, ALBERT \cite{Lan2020ALBERT} proposes a factorization of the embedding layer and a cross-layer parameter sharing as a way to improve model efficiency. Similar methodologies have also been previously explored by the Universal Transformer \cite{NIPS2019_8358}, and Deep Equilibrium Models \cite{dehghani2018universal}, demonstrating the effectiveness of weight-tied Transformers.

A second strategy is to distill the knowledge of pre-trained models into smaller models. For instance, PKD-BERT \cite{sun-etal-2019-patient}, TinyBERT \cite{jiao2020tinybert}, MobileBERT \cite{sun-etal-2020-mobilebert} and DistilBERT \cite{sanh2020distilbert} compress the knowledge of large models (teachers) into lighter ones (students). In this way, a compact model is obtained with a performance usually close to the original one.

While these approaches reduce the computational demand for execution, they are only available for the English language. In the same spirit, as pre-trained language models are increasingly common, it is essential to have these efficient alternatives available in other languages and thus alleviate excessive computational consumption.

\section{Data and Models}\label{sec:data}

\subsection{Data}

The data\footnote{\url{https://github.com/josecannete/spanish-corpora}} used to train all models was the same as that used to train BETO \cite{CaneteCFP2020}, which is an updated version of the dataset proposed by \newcite{cardellinoSBWCE}. This dataset has approximately 3 billion words which includes all Spanish Wikipedia and almost all the Spanish portion of the OPUS Project \cite{tiedemann2012parallel}.

\subsection{Spanish ALBERT (ALBETO)}

ALBERT \cite{Lan2020ALBERT} is a more efficient BERT-style model in terms of parameters because it uses the weight-tied strategy, which means to share all parameters across layers of the model.
We not only train a model comparable to BETO, but we also train larger models.
We introduce 5 ALBETO models: \textit{tiny}, \textit{base}, \textit{large}, \textit{xlarge} and \textit{xxlarge}. All except from the tiny one are similar in size to those trained by \newcite{Lan2020ALBERT}. ALBETO \textit{tiny} has the same size and configuration as Chinese ALBERT \textit{tiny} \footnote{\url{https://github.com/ckiplab/ckip-transformers}}. The complete configuration of every model can be seen on Table \ref{table:albeto-configurations}.

These five models share a vocabulary of 31K lowercase tokens, that was constructed using SentencePiece \cite{kudo2018sentencepiece} over the training dataset.
We trained all the ALBETO models using the LAMB optimizer \cite{you2019large} following all the guidelines used by the authors to obtain better results. We trained each model using a single TPU v3-8. The details about the training of each model can be seen in Table \ref{table:training-details-albetos}.

\begin{table}[]
\begin{center}
\resizebox{\columnwidth}{!}{
\begin{tabular}{|l|c|c|c|c|}
\hline
\textbf{Model}          & \textbf{Parameters} & \textbf{Layers} & \textbf{Hidden} & \textbf{Embedding} \\ \hline
ALBETO \textit{tiny}    & 5M         & 4      & 312    & 128       \\ \hline
ALBETO \textit{base}    & 12M        & 12     & 768    & 128       \\ \hline
ALBETO \textit{large}   & 18M        & 24     & 1024   & 128       \\ \hline
ALBETO \textit{xlarge}  & 59M        & 24     & 2048   & 128       \\ \hline
ALBETO \textit{xxlarge} & 223M       & 12     & 4096   & 128       \\ \hline
\end{tabular}}
\end{center}
\caption{The configurations of each ALBETO model trained on this work.}
\label{table:albeto-configurations}
\end{table}

\begin{table*}[]
\begin{center}
\resizebox{\textwidth}{!}{
\begin{tabular}{|l|c|c|c|c|c|c|}
\hline
\textbf{Model}          & \textbf{Learning Rate }  & \textbf{Batch Size} & \textbf{Warmup Ratio} & \textbf{Warmup Steps} & \textbf{Total Steps} & \textbf{Training Time (days)} \\ \hline
ALBETO \textit{tiny}    & 1.25e-3         & 2,048       & 1.25e-2       & 125,000       & 8,300,000     & 58.2                       \\ \hline
ALBETO \textit{base}    & 8.83e-4 & 960        & 6.25e-3      & 53,333        & 3,650,000     & 70.4                       \\ \hline
ALBETO \textit{large}   & 6.25e-4        & 512        & 3.12e-3     & 12,500        & 1,450,000     & 42.0                         \\ \hline
ALBETO \textit{xlarge}  & 3.12e-4       & 128        & 7.81e-4   & 6,250         & 2,775,000     & 64.2                       \\ \hline
ALBETO \textit{xxlarge} & 3.12e-4       & 128        & 7.81e-4   & 3,125         & 1,650,000     & 70.7                       \\ \hline
\end{tabular}}
\caption{Training details of all ALBETO models, which were trained using a single TPU v3-8 each one. }
\label{table:training-details-albetos}
\end{center}
\end{table*}

\subsection{Spanish DistilBERT (DistilBETO)}

We trained the DistilBETO model using the distillation technique to transfer the knowledge of the BETO model to this new model following the work of \newcite{sanh2020distilbert}. The architecture of DistilBETO is based on BETO but we removed the \textit{token-type embeddings}, the \textit{pooler layer} and decreased the number of layers from 12 to 6. Due to the limitations of time and hardware that do not allow us to test different architectures, we chose the one that obtained the best results according to the different ablation  experiments by \newcite{sanh2020distilbert}. 
DistilBETO was trained during 90k steps using a single GPU NVIDIA RTX 3090 following the best model training guidelines obtained by \newcite{sanh2020distilbert}.

\section{Evaluation}\label{sec:eval}

\begin{table}[]
\begin{center}
\begin{tabular}{|l|c|c|}
\hline
\textbf{Model}  & \textbf{POS}            & \textbf{NER}       \\ \hline
BETO \textit{uncased}   & 97.70           & 83.76     \\ \hline
BETO \textit{cased}     & \textbf{98.84} & \textbf{88.24}  \\ \hline
DistilBETO    & 97.50           & 81.19     \\ \hline
ALBETO \textit{tiny}    & 97.04          & 75.11     \\ \hline
ALBETO \textit{base}    & 98.08          & 83.35     \\ \hline
ALBETO \textit{large}   & 97.87          & 83.72     \\ \hline
ALBETO \textit{xlarge}  & 98.06          & 82.30      \\ \hline
ALBETO \textit{xxlarge} & 98.35          & 84.36     \\ \hline
\end{tabular}
\caption{Comparison of ALBETO, DistilBETO and BETO models on the test set for the tasks of POS and NER, which are sequence tagging tasks and are evaluated using F1 score as metric.}
\label{table:results-finetuning-ner-pos}
\end{center}
\end{table}

\begin{table}[]
\centering
\begin{tabular}{|l|c|c|c|}
\hline
\textbf{Model }  &    \textbf{MLDoc}       & \textbf{PAWS-X}       & \textbf{XNLI}           \\ \hline
BETO \textit{uncased}   &    96.38       & 84.25         & 77.76          \\ \hline
BETO \textit{cased}     &    96.65       & 89.80          & 81.98          \\ \hline
DistilBETO    &    96.35       & 75.80          & 76.59          \\ \hline
ALBETO \textit{tiny}    &    95.82       & 80.20          & 73.43          \\ \hline
ALBETO \textit{base}    &    96.07       & 87.95         & 79.88          \\ \hline
ALBETO \textit{large}   &    92.22       & 86.05         & 78.94          \\ \hline
ALBETO \textit{xlarge}  &    95.70        & 89.05         & 81.68          \\ \hline
ALBETO \textit{xxlarge} &\textbf{96.85}  & \textbf{89.85} & \textbf{82.42} \\ \hline
\end{tabular}
\caption{Comparison of ALBETO, DistilBETO and BETO models on the test set for the tasks of MLDoc, PAWS-X and XNLI. These tasks are treated as sentence classification tasks and use the accuracy as evaluation metric.}
\label{table:results-finetuning-mldoc-pawsx-xnli}
\end{table}

\begin{table*}[]
\begin{center}
\begin{tabular}{|l|c|c|c|}
\hline
\textbf{Model} & \textbf{MLQA}          & \textbf{SQAC}          & \textbf{TAR, XQuAD}    \\ \hline
BETO \textit{uncased}   & 64.12 / 40.83          & 72.22 / 53.45          & 74.81 / 54.62          \\ \hline
BETO \textit{cased}     & 67.65 / 43.38          & 78.65 / 60.94          & 77.81 / 56.97          \\ \hline
DistilBETO     & 57.97 / 35.50          & 64.41 / 45.34          & 66.97 / 46.55          \\ \hline
ALBETO \textit{tiny}    & 51.84 / 28.28          & 59.28 / 39.16          & 66.43 / 45.71          \\ \hline
ALBETO \textit{base}    & 66.12 / 41.10          & 77.71 / 59.84          & 77.18 / 57.05          \\ \hline
ALBETO \textit{large}   & 65.56 / 40.98          & 76.36 / 56.54          & 76.72 / 56.21          \\ \hline
ALBETO \textit{xlarge}  & 68.26 / 43.76          & 78.64 / 59.26          & \textbf{80.15 / 59.66} \\ \hline
ALBETO \textit{xxlarge} & \textbf{70.17 / 45.99} & \textbf{81.49 / 62.67} & 79.13 / 58.40          \\ \hline
\end{tabular}
\caption{Comparison of ALBETO, DistilBETO and BETO models on the task of QA. We show the results of the test set in each case. The task uses two metrics which are showed as F1 Score / Exact Match. }
\label{table:results-finetuning-qa}
\end{center}
\end{table*}

\begin{table*}[]
\begin{center}
\begin{tabular}{|l|c|c|c|c|}
\hline
\textbf{Model}          & \textbf{Parameters} & \textbf{Evaluation Average} & \textbf{Size}  & \textbf{Performance} \\ \hline
BETO \textit{uncased}   & 110M       & 77.48              & 1x    & 0.95x       \\ \hline
BETO \textit{cased}     & 110M       & 81.02               & 1x    & 1x          \\ \hline
DistilBETO    & 67M        & 73.22              & 1.64x & 0.90x       \\ \hline
ALBETO \textit{tiny}    & 5M         & 70.86              & 22x   & 0.87x       \\ \hline
ALBETO \textit{base}    & 12M        & 79.35              & 9.16x & 0.97x       \\ \hline
ALBETO \textit{large}   & 18M        & 78.12              & 6.11x & 0.96x       \\ \hline
ALBETO \textit{xlarge}  & 59M        & 80.20             & 1.86x & 0.98x       \\ \hline
ALBETO \textit{xxlarge} & 223M       & 81.34              & 0.49x & 1x       \\ \hline
\end{tabular}
\caption{Comparison of different models in terms of task performance and size. On the last two columns of the table we present a direct comparison of BETO (in particular the \textit{cased} model, which is the best of the two versions) with all the other models in terms of size (parameter count) and task performance (evaluation average of all tasks). }
\label{table:comparison-to-beto}
\end{center}
\end{table*}

\subsection{Tasks}

We evaluated all our models on 6 tasks, which are all part of GLUES \cite{CaneteCFP2020}, and we also closely follow the experimental setup proposed by them. We next describe the tasks.

\subsubsection{Document Classification}

Document classification is the task of assigning an entire document to an appropriate category. The categories depend on the chosen dataset and can range from topics. For this task we are using the Spanish part of MLDoc corpus \cite{SCHWENK18.658} which uses a subset of the Reuters Corpus \cite{lewis2004rcv1}.
    
\subsubsection{Part of Speech}

Part-of-speech tagging (POS tagging) is a sequence labeling task that consists of tagging words in a text with their corresponding syntactic categories or part-of-speech. A part of speech is a category of words with similar grammatical properties. Common parts of speech are noun, verb, adjective, adverb, pronoun, preposition, conjunction, etc. The dataset used for this task is the Spanish subset of Universal Dependencies (v1.4) Treebank \cite{udv1.4}. 

\subsubsection{Named Entity Recognition}

Named Entity Recognition (NER) is another sequence labeling task, in which one tries to label entities in the text with their corresponding type, which can be names of people, organizations, places and miscellaneous items. Approaches typically use the BIO notation, which differentiates the beginning (B) and the inside (I) of entities. O is used for tokens that are not entities.
For Named-Entity Recognition we are using the Spanish part of the Shared Task of CoNLL-2002 \cite{sang2002introduction}.

\subsubsection{Paraphrase Identification}

The task of Paraphrase Identification consists of verifying whether two sentences are semantically equivalent or not. We are using PAWS-X \cite{yang2019pawsx} dataset, which is a multilingual version of PAWS \cite{zhang2019paws} dataset. The dataset provides a machine translated train set and professionally translated development and test sets for different languages. We used the Spanish portion of it.

\subsubsection{Natural Language Inference}

Natural language inference is the task of determining whether a ``hypothesis'' is true (entailment), false (contradiction), or undetermined (neutral) given a ``premise''.
For Natural Language Inference we are using the Spanish part of XNLI \cite{conneau2018xnli} which has a training set that is a machine translation of MultiNLI \cite{williams2017broad} and development and test set that were professional translated.

\subsubsection{Question Answering}

The task of Question Answering consists of, given a context and a question about that context, highlighting the sequence of words within that context that answers the question. This task uses two metrics to evaluate performance, which are F1 Score and Exact Match. For this task we considered four different datasets: MLQA \cite{lewis2019mlqa}, TAR \cite{carrino2019automatic}, XQuAD \cite{Artetxe:etal:2019} and SQAC \cite{gutierrez2022maria}. MLQA and TAR datasets offer a different machine translated version of SQUaD v1.1 \cite{rajpurkar2016squad} to Spanish with train and development sets. Also, MLQA offers a development set and test set professionally translated to Spanish. XQuAD offers a different professionally translated test set for Spanish. SQAC is notably the only one which was created exclusively for Spanish and offers train, development and test sets. In our experiments the machine translated development set of MLQA is left out in favor of the professionally translated set. And in the case of XQuAD which only offers test set, we used TAR as train and development sets.

\subsection{Fine-tuning}

In order to have a fair comparison, we fine-tuned all our models as well as BETO in both versions, \textit{uncased} and \textit{cased}, using the same code for all of them, which uses PyTorch \cite{paszke2019pytorch} and the HuggingFace's Transfomers library \cite{wolf2019huggingface}. We follow the procedure described next.

We fine-tuned our models using the standard way proposed by \cite{devlin-etal-2019-bert}. The only preprocessing performed is tokenization according to the token vocabulary of each model which converted words into subwords and added the special tokens [CLS], [PAD] and [SEP] to each sentence. We set the maximum length of an input sentence to 512 tokens in all models and sentences were truncated to it when larger.

We did a hyperparameter search using combinations of batch size \{16, 32, 64\}, learning rate \{1e-5, 2e-5, 3e-5, 5e-5\} and number of epochs \{2, 3, 4\} on BETO, DistilBETO, ALBETO \textit{tiny} and \textit{base}.

For the larger models (ALBETO \textit{large}, \textit{xlarge} and \textit{xxlarge}), due to numerical instability in the training process we used smaller learning rates \{1e-6, 2e-6, 3e-6, 5e-6\}. 

We used between one and two GPUs NVIDIA RTX 3090 for fine-tuning depending on the model and task. To fine-tune the largest models on QA we used two GPUs NVIDIA A100 from the Patagón supercomputer \cite{patagon-uach}. In some cases the memory of the GPUs was not enough so we used gradient accumulation to reach the desired batch size.
We then selected the models with best results on the development set. 

\subsection{Results}

Table \ref{table:results-finetuning-ner-pos} shows the results of all models evaluated on POS and NER tasks in terms of F1 score. 

It can be seen that for the POS, BETO \textit{cased} surpasses ALBETO \textit{xxlarge}, the second-best, by a 0.4 percentual difference (pd). Similar behavior is observed for the NER task where the BETO \textit{cased} model outperforms ALBETO \textit{xxlarge} by a considerable margin of 4.5 pd. This can be explained by the type of task since identifying entities that usually start with capital letters allows the \textit{cased} model to have additional knowledge to solve this task, compared to other models. 

Concerning the results of the performance of the models on MLDoc, PAWS-X and XLNI tasks in terms of accuracy are shown in Table \ref{table:results-finetuning-mldoc-pawsx-xnli}. It is observed that, in general, the largest ALBETO models (ALBETO \textit{xxlarge}) outperform other models by a small margin. This behavior can be explained since this model has more capacity than the other presented models. In the MLDoc task, ALBETO \textit{xxlarge} surpasses BETO \textit{cased} by 0.2 pd. Regarding the PAWS-X task, ALBETO \textit{xxlarge} also outperforms the BETO \textit{cased} model by a small margin of 0.05 pd. Finally, for XNLI, ALBETO \textit{xxlarge} achieves the best performance with an F1 score of 82.42, surpassing BETO \textit{cased}, the second-best, by 0.5 pd.  

The results on the task of Question Answering are showed in Table  \ref{table:results-finetuning-qa}. We can observe that generally the models on this task follow the same behaviour of that in the sentence classification tasks, with ALBETO \textit{xxlarge} or ALBETO \textit{xlarge} outperforming BETO by a considerable margin. In the MLQA dataset, ALBETO \textit{xxlarge} surpasses all other models, with a difference of 2.5 pd in F1 score and 2.6 pd in exact match, respect to BETO \textit{cased}, the third-best. In the SQAC dataset, ALBETO \textit{xxlarge} outperforms BETO \textit{cased} by 2.8 pd in F1 score and 1.7 pd in exact match. In the case of XQuAD (trained on TAR), ALBETO \textit{xlarge} is the best one, followed by ALBETO \textit{xxlarge} and BETO \textit{cased}, with a difference of 2.3 pd in F1 score and 2.7 pd in exact match between ALBETO \textit{xlarge} and BETO \textit{cased}.




\subsection{Model Size and Task Performance}

Table \ref{table:comparison-to-beto} shows a comparison of all models in terms of model size and overall task performance. We average the results of each dataset to have a score of each model in the same way of the GLUE \cite{wang2018glue} benchmark does. For the Question Answering task, where we have two different metrics, we first average them to have a single score for every dataset. We can notice that in terms of parameters, ALBETO \textit{tiny} is the smaller model on the comparison while ALBETO \textit{xxlarge} being the largest one. In terms of overall task performance ALBETO \textit{xxlarge} and BETO \textit{cased} are the most competitive ones.

The last two columns of the table also compare all the models with the size and performance of BETO \textit{cased}. Some of the most promising results are ALBETO \textit{tiny}, which is able to retain 87\% of BETO performance while being 22 times smaller in size and ALBETO \textit{base}, which has 97\% of BETO performance while having more than 9 times less parameters.

\section{Conclusion}\label{sec:conc}

This paper presents DistilBETO and five ALBETO models (\textit{tiny}, \textit{base}, \textit{large}, \textit{xlarge}, and \textit{xxlarge}), comprising six new pre-trained language models based on the ALBERT and DistilBERT architectures for the Spanish language. While the ALBETO models exploit the weight-tied strategy to be more parameter efficient, DistilBETO was built by compressing  BETO into a lighter model using a knowledge distillation approach. We also comprehensively evaluated each proposed model fine-tuned on a set of NLP tasks for Spanish (POS, NER, MLDOC, PAWS-X, XNLI, MLQA, SQAC and XQuAD). Our results indicate that the proposed models are competitive with the current models available for Spanish and are much more efficient in their number of parameters. 

We hope this work will expand the availability of pre-trained language models based on the Spanish language to gather a wider NLP community, including researchers, developers, and students.  

We envision several avenues of future research. First, we expect to evaluate these models on more tasks to increase the coverage of GLUES, which is our current evaluation benchmark \cite{CaneteCFP2020}. In addition, we want to explore whether the modifications made to the DistilBETO architecture according to \cite{sanh2020distilbert} are the best for Spanish models.
Also, we also want to further analyze the fine-tuning and inference speed of these models. In addition, we plan to release more distilled models fine-tuned explicitly for many NLP tasks. Finally, we intend to study the capabilities of these resources in few-shot and zero-shot learning settings.

\section{Acknowledgements}
This work was supported in part by the National Center for Artificial Intelligence CENIA FB210017, Basal ANID and by the Patagón supercomputer of Universidad Austral de Chile (FONDEQUIP EQM180042). Felipe Bravo-Marquez was supported by ANID FONDECYT grant 11200290, U-Inicia VID Project UI-004/20 and ANID -Millennium Science Initiative Program - Code ICN17\_002.

\section{Bibliographical References}\label{reference}

\bibliographystyle{lrec2022-bib}
\bibliography{lrec2022-references}

\end{document}